\documentclass[10pt,twocolumn,letterpaper]{article}

\usepackage{iccv}
\usepackage{times}
\usepackage{epsfig}
\usepackage{graphicx}
\usepackage{amsmath}
\usepackage{amssymb}
\usepackage{color}
\usepackage{xcolor}
\usepackage{arydshln} 
\usepackage{hhline}
\usepackage{amssymb}
\usepackage{pifont}
\usepackage{enumitem}
\usepackage{colortbl}
\newcommand{\cmark}{\ding{51}}%
\definecolor{citecolor}{HTML}{0071bc}

\definecolor{Gray}{gray}{0.5}

\usepackage[pagebackref=true,breaklinks=true,letterpaper=true,colorlinks,citecolor=citecolor,bookmarks=false]{hyperref}

\iccvfinalcopy 

\def\iccvPaperID{8399} 

\newlength\savewidth\newcommand\shline{\noalign{\global\savewidth\arrayrulewidth
  \global\arrayrulewidth 1pt}\hline\noalign{\global\arrayrulewidth\savewidth}}
\newcommand{\tablestyle}[2]{\setlength{\tabcolsep}{#1}\renewcommand{\arraystretch}{#2}\centering\footnotesize}
\makeatletter\renewcommand\paragraph{\@startsection{paragraph}{4}{\z@}
  {.5em \@plus1ex \@minus.2ex}{-.5em}{\normalfont\normalsize\bfseries}}\makeatother

\ificcvfinal\fi

\begin{document}

\title{Rethinking Self-supervised Correspondence Learning: \\ A Video Frame-level Similarity Perspective}

\author{Jiarui Xu \qquad Xiaolong Wang\\
UC San Diego 
}

\maketitle
\ificcvfinal\thispagestyle{empty}\fi

\begin{abstract}

Learning a good representation for space-time correspondence is the key for various computer vision tasks, including tracking object bounding boxes and performing video object pixel segmentation. To learn generalizable representation for correspondence in large-scale, a variety of self-supervised pretext tasks are proposed to explicitly perform object-level or patch-level similarity learning. Instead of following the previous literature, we propose to learn correspondence using Video Frame-level Similarity (VFS) learning, i.e, simply learning from comparing video frames. Our work is inspired by the recent success in image-level contrastive learning and similarity learning for visual recognition. Our hypothesis is that if the representation is good for recognition, it requires the convolutional features to find correspondence between similar objects or parts. Our experiments show surprising results that VFS surpasses state-of-the-art self-supervised approaches for both OTB visual object tracking and DAVIS video object segmentation. We perform detailed analysis on what matters in VFS and reveals new properties on image and frame level similarity learning. Project page with code is available at \href{https://jerryxu.net/VFS}{https://jerryxu.net/VFS}.

\end{abstract}

\section{Introduction}

Learning visual correspondence across space and time is one of the most fundamental problems in computer vision. It is widely applied in 3D reconstruction, scene understanding, and modeling object dynamics. The research of learning correspondence in videos can be cast into two categories: the first one is learning object-level correspondence for visual object tracking~\cite{sethi_finding_1987,ramanan2005strike,valmadre2017end}, relocalizing the object with bounding boxes along the video; the other one is learning fine-grained correspondence, which is commonly applied in optical flow estimation~\cite{horn1981determining,fischer2015flownet} and video object  segmentation~\cite{Cae17,yang2018efficient}. While both lines of research have been extensively explored, most approaches acquire training supervision from simulations or limited human annotations, which increases the difficulty for generalization across different data and tasks. 

\begin{figure}[t]
\centering
\includegraphics[width=.99\linewidth]{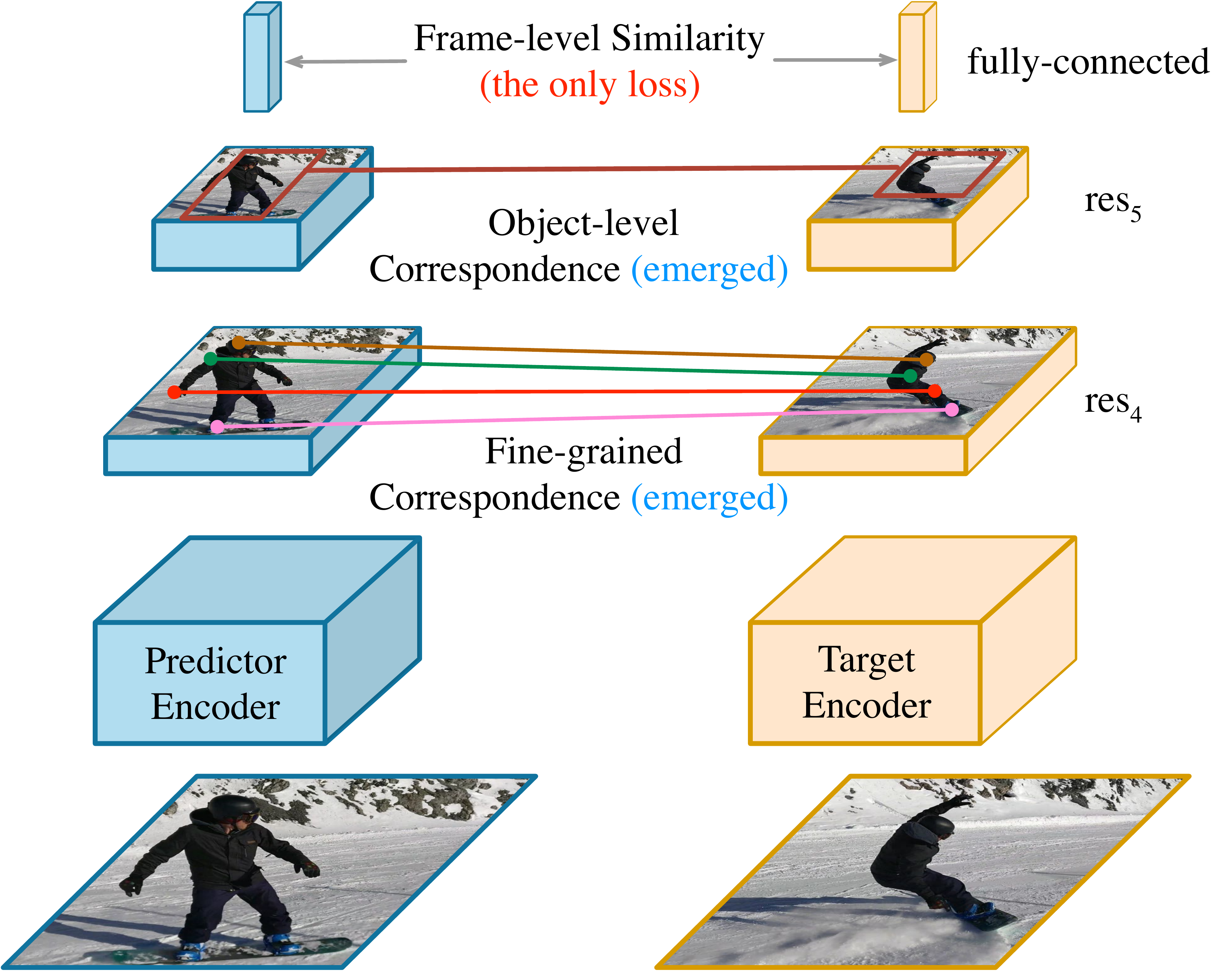}
\vspace{.8em}
\caption{
Video Frame-level Similarity (VFS) learning. It compares the fully-connected layer embeddings of frames from the same video for learning.
By minimizing the frame-level feature distance,
the fine-grained and object-level correspondence can automatically emerge in res$_4$ and res$_5$ blocks in the ResNet architecture, without using any explicit tracking-based pretext task. 
\label{fig:teaser} 
\vspace{-.8em}
}
\end{figure}

One way to tackle this problem is to learn representations for correspondence using free temporal supervision signals in videos. Recently, a lot of efforts have been made in self-supervised learning of space-time visual correspondence~\cite{carl2018color,wang2019cycle,jabri2020walk}, where different pretext tasks are designed to learn to track pixels or objects in videos. For example, Wang et al.~\cite{wang2019cycle} propose to use the cycle-consistency of time (i.e., forward-backward object tracking) as a supervisory signal for learning. Building on this, Jabri et al.~\cite{jabri2020walk} combine the cycle-consistency of time with patch-level similarity learning and achieve a significant improvement in learning correspondence. Given this encouraging result, we take a step back and ask the questions: Do we really need to design self-supervised object (or patch) tracking task explicitly to learn correspondence? Can image-level similarity learning alone learn the correspondence?

Recent development in image-level similarity learning (e.g., contrastive learning) has shown the self-supervised representation can be applied to different downstream semantic recognition tasks, and even surpassing the ImageNet pre-training networks~\cite{he2020moco,chen2020simclr, chen2020mocov2,caron2020swav, misra2020pirl, wu2018nipd, henaff2020cpcv2, tian2019cmc, tian2020infomin}. Our hypothesis is that if the higher-level fully-connected layer feature encodes the object structure and semantic information, it needs to be supported by the ability of finding correspondence between similar object instances and between object parts~\cite{zeiler2014visualizing}. This forces the convolutional  representation to implicitly learn visual correspondence. 

With this hypothesis, we propose to perform Video Frame-level Similarity (VFS) learning for space-time correspondence without any explicit tracking-based pretext task. As illustrated in Figure~\ref{fig:teaser}, we forward one pair or multiple pairs of frames from the same video into a siamese network, and compute the similarity between the frame-level features (fully-connected layer embeddings) for learning the network representation. We examine the learning with negative pairs as~\cite{chen2020simclr,he2020moco} and without negative pairs as~\cite{grill2020bootstrap,chen2020simsiam} under our VFS framework. We build our model based on the ResNet architecture~\cite{he2016resnet}. During inference, we use the res$_4$ features for fine-grained correspondence task (e.g., DAVIS object segmentation~\cite{jordi2017davis}) and the res$_5$ features for object-level correspondence task (e.g., OTB object tracking~\cite{yi2015otb}). Surprisingly, we find VFS can surpass state-of-the-art self-supervised correspondence learning approaches~\cite{jabri2020walk,li2019uvc}. Based on our experiments, we observe the following key elements for VFS:

(i) Training with large frame gaps and multiple frame pairs improves correspondence. When sampling a pair of frames from the same video for similarity learning, we observe that increasing the time differences between the two frames can improve fine-grained correspondence noticeably on DAVIS ($\sim 3\%$), and object-level correspondence significantly on OTB ($> 10\%$). Training with multiple frame pairs at the same time achieves further improvement.

(ii) Training with color augmentation is harmful for fine-grained  correspondence, but beneficial for object-level correspondence. With color augmentation, the performance on the DAVIS dataset is decreased ($\sim 3\%$) while it significantly improved performance  on the OTB dataset ($\sim 10\%$), which indicates the feature learns better object invariance.

(iii) Training without negative pairs improves both fine-grained and object-level correspondences. Recent literature has shown that similarity learning for visual representation is achievable even without negative pairs~\cite{grill2020bootstrap,chen2020simsiam}. While the results are surprising, it was still unclear how it can be beneficial for performance. In this paper, we show that VFS without negative training pairs can improve representations for different levels of correspondence.

(iv) Training with deeper networks gives significant improvements. While deeper networks generally improves recognition performance, it is not the case when training with self-supervised tracking pretext tasks for correspondence. We observe very small improvement or even worse correspondence results when using ResNet-50 compared to ResNet-18 with previous approaches~\cite{wang2019cycle,jabri2020walk,li2019uvc}. When learning correspondence with VFS implicitly, we achieve much better performance when using a deeper model.

Given our detailed analysis and state-of-the-art performance, we hope VFS can serve as a strong baseline for self-supervised correspondence learning. The study of intermediate representations for correspondence also provides a better understanding on what image-level  self-supervised similarity learning has learned. Finally, VFS also reveals the new property of similarity learning without negative pairs: it improves both object-level and fine-grained correspondence.

\section{Related Work}

\paragraph{Temporal Correspondence.} Learning correspondence from video frames is a long stand problem in computer vision. We can classify the temporal correspondence into two categories. The first one is the  \textit{fine-grained} correspondence which has been widely studied in optical flow and motion estimation~\cite{lucas1981iterative, horn1981determining, memin1998dense, brox2004high, brox2009large, sun2010secrets, liu2011sift}. For example, Brox and Malik~\cite{brox2009large} proposed a region hierarchy matching approach to perform dense and long-range flow estimation. Recently, deep learning based approaches have been applied to estimate optical flows by training on synthetic datasets~\cite{butler2012naturalistic, dosovitskiy2015flownet, ranjan2017optical, ilg2017flownet, sun2018pwc}. While largely improving the efficiency, training on synthetic data largely restricts the network's generalization ability to real world scenes. The second one is finding \textit{object-level} correspondence which is meant to offer reliable and long-range visual object tracking~\cite{sethi_finding_1987, ramanan2005strike, yang2005efficient, wu_situ_2007, pan2009recurrent, kalal2010forward}. While tracking by training a detector to perform per-frame recognition offers promising results~\cite{ramanan2005strike, andriluka2008people, kalal2012tracking, wang2013learning, bergmann2019tracktor, li2018high}, there is recent rise back to the classic tracking-by-matching methods~\cite{dorin2000meanshift, yang2005efficient, henriques2014high, li2014scale} using deep features~\cite{bertinetto2016siamfc,valmadre2017end}. For example, Bertinetto et al.~\cite{bertinetto2016siamfc} propose a fully-convolutional siamese network and adopt similarity learning for tracking. However, these approaches are still heavily relying on human annotations for training. In this paper, we propose a self-supervised similarity learning approach which learns both fine-grained and object-level correspondence.

\paragraph{Self-supervised Learning from Videos.} Self-supervised learning offers a way to learn generalizable visual representations with different pretext tasks~\cite{dosovitskiy2015discriminative,doersch2015unsupervised,pathak2016context,noroozi2016unsupervised,zhang2016colorful,gidaris2018unsupervised}. Beyond static images, temporal information from videos also offers rich supervision for representation learning~\cite{Goroshin2015,Agrawal2015,Jayaraman2015,Mathieu2015,Srivastava2015LSTMs,wang2015unsupervised,Li2016,Pathak2017,gordon2020vince, purushwalkam2020demystifying}. For example, Wang et al.~\cite{wang2015unsupervised} use off-the-shelf tracker to track objects in videos to provide supervisory signals. The learned representation has shown to be useful for multiple downstream recognition and geometry estimation tasks. Instead of learning a general representation, there is a line of recent research on self-supervised learning specifically for finding correspondence~\cite{carl2018color,wang2019cycle,wang2019udt,lai2019corrflow,lai2020mast,li2019uvc,jabri2020walk, purushwalkam2020aligning}. For example, Vondrick et al.~\cite{carl2018color} propose to propagate the current frame color to predict the future frame color as a pretext task to learn fine-grained correspondence between the current and future frame. Other tasks including tracking objects~\cite{wang2019cycle,wang2019udt} and patches~\cite{jabri2020walk,purushwalkam2020aligning} are also designed to explicitly find different levels of correspondences. But is explicit tracking task the only way to learn correspondence? Our work introduces an alternative perspective to learn correspondence implicitly via image-level similarity learning.

\paragraph{Image-level Similarity Learning.} The image-level similarity learning provides a way for visual representation learning. It uses a siamese network to enforce two different views or augmentations of the same image to have similar features. The recent proposed self-supervised contrastive learning is a one version of similarity learning~\cite{wu2018nipd, oord2018cpc, henaff2020cpcv2,ye2019unsupervised,bachman2019amdim,tian2019cmc,he2020moco,chen2020mocov2, misra2020pirl,chen2020simclr,caron2020swav,gordon2020vince,purushwalkam2020demystifying, cheng2021equivariant}. The idea is to learn representations via attracting the similar (positive) image pairs and repulsing a large number of dissimilar (negative) image pairs. For example, He et al.~\cite{he2020moco} propose to use a momentum  network to encode the large number of negatives for efficient contrastive learning. They have shown the learned representation achieves state-of-the-art performance when transferred to multiple downstream recognition tasks. Purushwalkam and Gupta~\cite{purushwalkam2020demystifying} further extend contrastive learning in temporal domain and show using temporal augmentation can improve object viewpoint invariance. Recently, it has been shown that the negative pairs are not necessary to learn a good visual representation~\cite{grill2020byol,chen2020simsiam}. While it  provides a better understanding for similarity learning, it is unclear if learning without negatives gives better representations. In this paper, we extend image-level similarity learning to video frames. Instead of performing recognition tasks, our aim is to learn visual correspondence. We also reveals that learning without negatives is beneficial for finding space-time correspondence.

\section{Method}

We propose Video Frame-level Similarity (VFS) learning for different levels of space-time correspondence. In this section, we will first introduce two common practices for image-level similarity learning for visual representations. Then we will unify these two paradigms under VFS.

\subsection{Background: Image-level Similarity Learning}
\label{sec:background}

The image-level similarity learning~\cite{wu2018nipd, he2020moco,grill2020byol, chen2020simclr,chen2020simsiam} provides a way to learn visual representations in a self-supervised manner. It learns the representation by minimizing the distance between two different augmented views of the same image in feature space. We classify the similarity learning into two types based on whether it is using the negative examples during training.

\paragraph{Similarity Learning with Negatives.} Contrastive learning~\cite{wu2018nipd,he2020moco,chen2020simclr} is similarity learning with negative sample pairs. In image-level contrastive learning, the positive pairs are different augmented views of the same image, and the negative pairs are from different images. The training objective is to push the representations of the positive (similar) pairs to be close to each other, while keep the representations of the negative (dissimilar) pairs to be far.

Formally, we denote $x$ and $x'$ as two different augmented views of an input image. The contrastive learning utilizes a siamese network architecture with a predictor encoder $\mathcal{P}$ and a target encoder $\mathcal{T}$. We use these two networks to extract features for both inputs respectively, and normalize the outputs with  $l_2$-normalization. The output embeddings for $x$ and $x'$ can be represented as $p \triangleq \mathcal{P}(x)/\|\mathcal{P}(x)\|_2$ and $z \triangleq \mathcal{T}(x')/\|\mathcal{T}(x')\|_2$. Let $\mathcal{U}=\{u_1, u_2, \dots, u_K\}$ be the negative bank that stores the features of negative samples.
The optimization objective is minimizing InfoNCE loss~\cite{oord2018cpc}, defined as 
\begin{equation}
\mathcal{L}_{p, z, \mathcal{U}} = -\log \frac{\exp (p{\cdot}z/\tau)}{\exp (p{\cdot}z/\tau) + \sum_{k=1}^K \exp(p{\cdot}u_k/\tau)}
\label{eq:infonce_img}
\end{equation} where $\tau$ is a temperature hyper-parameter~\cite{wu2018nipd}.

There are multiple ways to construct the negative sample bank including directly storing all the features~\cite{wu2018nipd} or sampling the negatives online~\cite{chen2020simclr}. In this paper, we adopt a recent practice proposed by He et al.~\cite{he2020moco}, which uses a momentum updated queue as the negative bank. To achieve this, the target encoder network $\mathcal{T}$ is updated as a moving average of the predictor encoder $\mathcal{P}$ as,
\begin{equation}
\xi \leftarrow  m\xi+ (1-m)\theta, ~m\in [0, 1),
\label{eq:momentum}
\end{equation} where $\theta, \xi$ are parameters of $\mathcal{P}$ and $\mathcal{T}$. Thus the encoders  $\mathcal{P}$ and $\mathcal{T}$ share the same network architecture in~\cite{he2020moco}.

\paragraph{Similarity Learning without Negatives.} Recently, researchers discover that similarity learning without negative sample pairs can achieve comparable performance as contrastive learning in visual representation learning~\cite{grill2020byol,chen2020simsiam}. Without the negative samples, the optimization objective can be simplified as the minimizing the cosine feature distance between two views
\begin{equation}
\mathcal{L}_{p, z}= \|p - z\|_2^2 = 2 - 2{\cdot}p{\cdot}z. 
\label{eq:cosine_img}
\end{equation}
However, optimizing this objecitve will easily lead to degenerated solution and yields a collapsed representation~\cite{grill2020byol,chen2020simsiam}. To resolve this issue, Chen et al.~\cite{chen2020simsiam} propose two techniques: (i) Share most of the parameters between the predictor encoder $\mathcal{P}$ and the target encoder  $\mathcal{T}$, except adding one additional MLP for the predictor encoder; (ii) Stop the gradients back-propagated from the loss to the target network. In this paper, we adopt this approach for similarity learning without negative data. Please refer to~\cite{chen2020simsiam} for more details.

\subsection{Video Frame-level Similarity Learning}
\label{sec:vfs}

\begin{figure}[t]
\centering
\includegraphics[width=.99\linewidth]{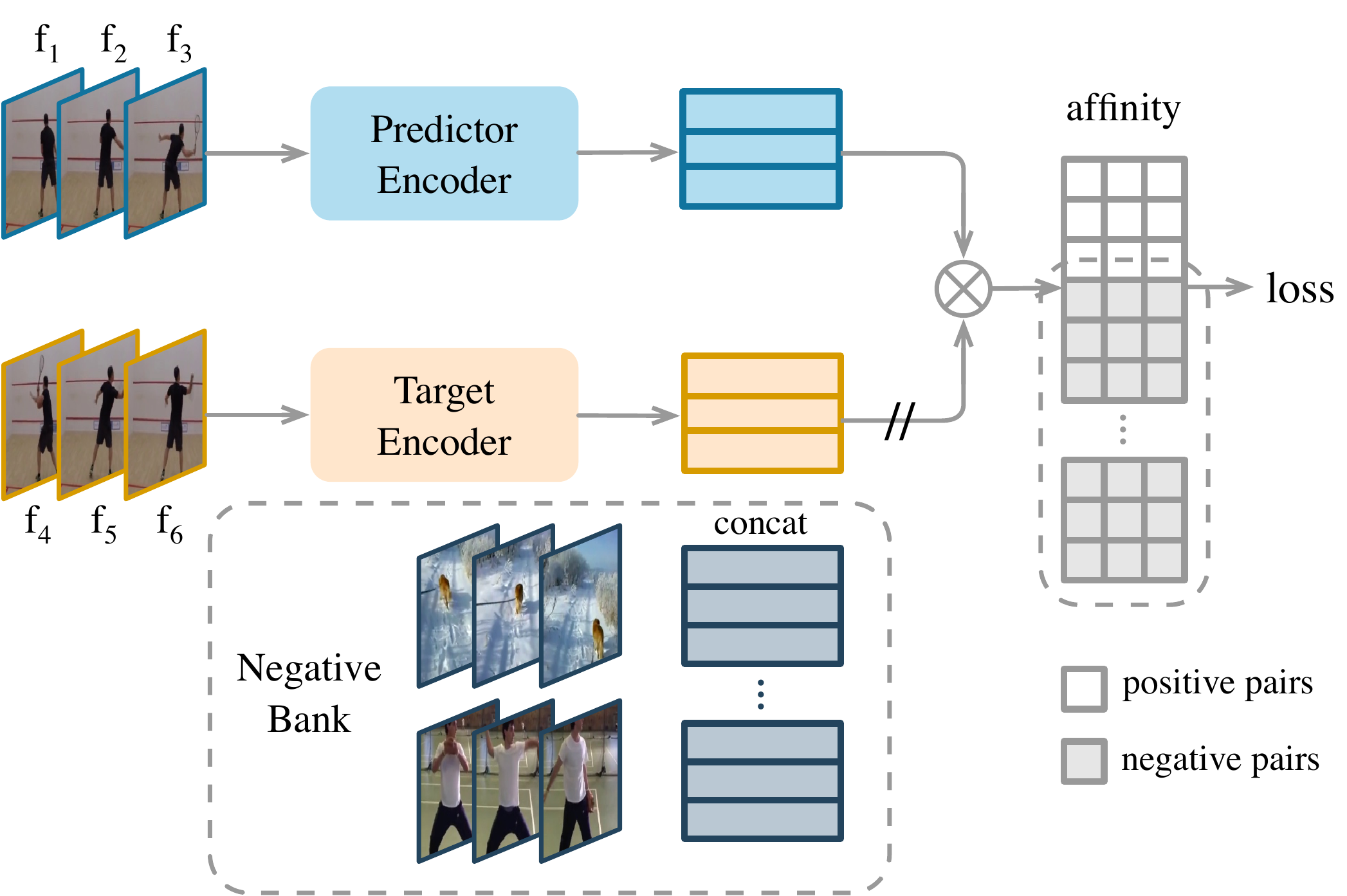}
\vspace{.5em}
\caption{
\textbf{Video Frame-level Similarity Pipeline}. 
The affinity matrix is the pairwise feature similarity between predictor features and target features.
The dash rectangle areas indicate the \textit{with negative pairs} case. 
The features of negative samples are stored in the negative bank and concatenate with target features.
The encoder is trained to maximize the affinity of positive pairs and minimize the affinity of negative ones.
\label{fig:pipeline}
\vspace{-0.8em}
}
\end{figure}

Building on image-level similarity learning, we propose to perform similarity learning between video frames, i.e., Video Frame-level Similarity learning (VFS). Our approach considers frames at different timestamps as different views for similarity learning. In a video clip with length $L$, there are $\frac{L^2}{2}$ possible positive sample pairs for learning. In VFS, each video frame is pulled towards a global video feature, resulting in a representation invariant to natural object deformation and viewpoint changes over time. In our experiments, we show that this learning objective can enforce the emergence of visual correspondence from the convolutional layers. Our hypothesis is that, like applying other augmentations in image-level similarity learning, sampling different temporal views also help the representation to learn object structure and even semantic information. This implicitly requires the convolutional features to learn about object and fine-grained correspondence. We will introduce our approach in details as follows.

\paragraph{Learning Objectives.} Given a video with $L$ frames $\{f_1, f_2, \dots, f_L\}$, we sample two random frames $f_i, f_j$ and apply data augmentation on them. We then forward both frames to the predictor encoder $\mathcal{P}$ and target encoder $\mathcal{T}$ to extract their features as $p_i \triangleq \mathcal{P}(f_i)/\|\mathcal{P}(f_i)\|_2$,  $z_j \triangleq \mathcal{T}(f_j)/\|\mathcal{T}(f_j)\|_2$. We provide two options for VFS learning with and without the negative pairs. Following Eq.~\ref{eq:infonce_img}, the objective for VFS \textit{with negative pairs} is represented as,
\begin{equation}
\mathcal{L}_{p_i, z_j, \mathcal{U}} = -\log \frac{\exp (p_i{\cdot}z_j/\tau)}{\exp (p_i{\cdot}z_j/\tau) + \sum_{k=1}^K \exp(p_i{\cdot}u_k/\tau)}. 
\label{eq:infonce_vid}
\end{equation}
On the other hand, VFS can also be trained \textit{without negative pairs} by following Eq.~\ref{eq:cosine_img}, and the objective is,
\begin{equation}
\mathcal{L}_{p_i, z_j}= \|p_i - z_j\|_2^2 = 2 - 2{\cdot}p_i{\cdot}z_j. 
\label{eq:cosine_vid}
\end{equation}
While we unify learning with and without negatives under VFS, we adopt the implementation details in~\cite{chen2020mocov2,chen2020simsiam} to adjust the architectures and training schemes accordingly as introduced in Section~\ref{sec:background}.

We illustrate our learning pipeline as Figure~\ref{fig:pipeline}. Beyond sampling one pair of data, learning from videos allows to sample multi-paired positive samples. To be more specific, we can sample $n$ frames from the videos, and divide them into two splits for predictor and target encoders. We then compute the feature similarity between $\frac{n}{2}$ predictor features and $\frac{n}{2}$ target features which yields a affinity matrix is of shape $\frac{n}{2} \times \frac{n}{2}$. When training without negatives, we can apply Eq.~\ref{eq:cosine_vid} to maximize each element in the affinity matrix, i.e., minimize the distance between each video frame pair. When training with $K$ negative examples, the shape of the affinity matrix becomes $\frac{n}{2}\times(\frac{n}{2}+K)$. We apply Eq.~\ref{eq:infonce_vid}  for learning, where the negative bank $\mathcal{U}=\{u_1, u_2, \dots, u_K\}$ are sampled from other videos. We  illustrate the sampling method for the $n$ frames as follows.

\begin{table}[t]
\centering
\small
\tablestyle{3pt}{1.1}
\begin{tabular}{l|c|c}
method    & $I_i$  & $\mathbb{E}(I_i-I_{i-1})$ \\
\shline
Continuous Sampling & $I_1 + (i-1)\delta$ & $\delta$ \\
Distant Sampling    & $\frac{L}{n}(i-1) + \mathrm{unif}(0, \frac{L}{n})$ & $\frac{L}{n}$
\end{tabular}
\vspace{.8em}
\caption{
\textbf{Video Frame Sampling}. 
$I_i$ is the i-th sampled frame index.
$\delta$ is the frame interval of \textit{continuous sampling}. 
$n$ is the number of segments (frames) of \textit{distant sampling}.
\label{table:sampling}
\vspace{-.5em}
}
\end{table}

\begin{figure}[t]
\centering
\includegraphics[width=.9\linewidth]{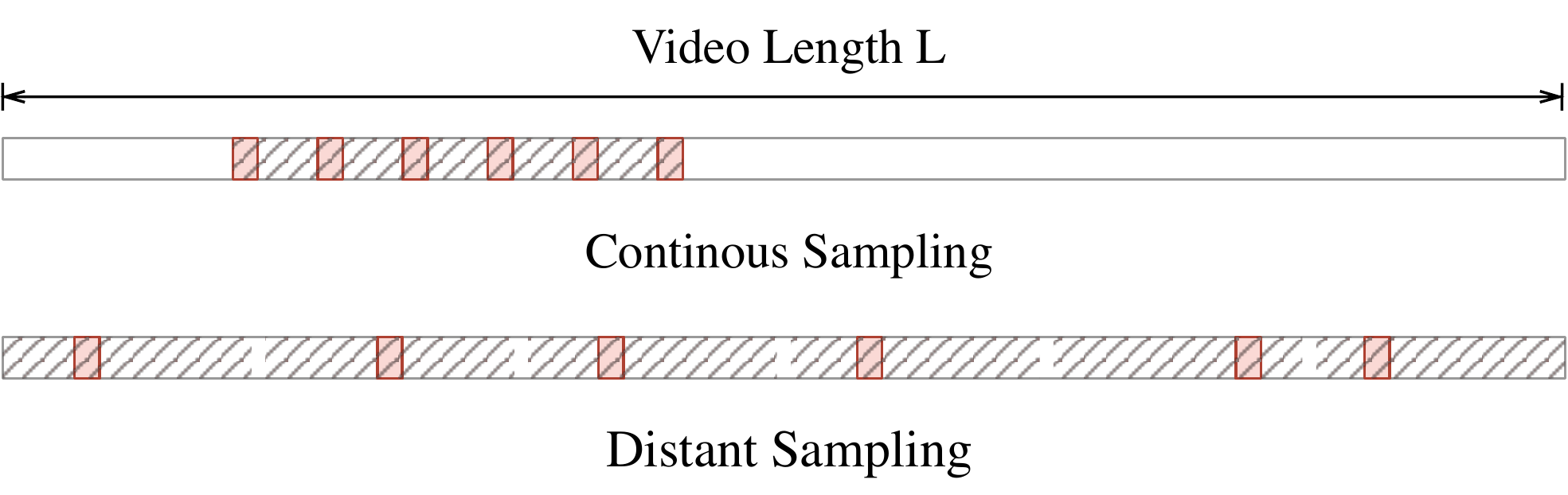}
\vspace{.2em}
\caption{
\textbf{Video Frame Sampling}. 
Dash areas are selected segments for sampling.
Solid blocks indicate sampled frames. 
\textit{Continuous sampling} yields temporally continuous frames, while \textit{distant sampling} samples frames with the larger displacement as well as randomness.
\label{fig:sampling}
\vspace{-1em}
}
\end{figure}

\paragraph{Temporal Sampling.} 

Given a video with $L$ frames, sampling $n$ frames yields a set of indices $\{I_1, I_2, \dots, I_n\}$, where $I_i \in [1, L]$. In this paper, we investigate two strategies for temporal sampling: (i) \textit{Continuous sampling}, which first selects a starting frame index and then continuously samples with a fixed frame interval $\delta$; (ii) \textit{Distant sampling}, which splits the video into $n$ disjoint segments, then randomly selects 1 frame from each segment as $I_i = \frac{L}{n}(i-1) + \mathrm{unif}(0, \frac{L}{n})$. We illustrate the two sampling strategies in both Table~\ref{table:sampling} and Figure~\ref{fig:sampling}. The continuous sampling strategy is widely applied in architectures that exploits 3D Convolution~\cite{tran2015learning, carreira2017quo} for  locally consistent feature map~\cite{xie2018rethinking}. On the other hand, the distant sampling provides a larger coverage of the entire video and a more aggressive augmentation effect. We will ablate how the sampling strategy affects correspondence learning in our experiments. 

\paragraph{Data Augmentation.} 

Besides using temporal signals to provide different views of training data, we also adopt the common data augmentation practices in image-level similarity learning~\cite{wu2018nipd,chen2020mocov2, chen2020simclr}. We apply the spatial augmentation (e.g., random cropping and flipping) and color augmentation (e.g., grayscale and color jitter) in our experiments. Specifically, we observe the color augmentation plays an important role in correspondence learning and we will report our findings in the experiment section.

\section{Experiments}

We perform experiments on representation learned by VFS for fine-grained correspondence tasks~\cite{jordi2017davis,jhuang2013jhmdb,zhou2018vip} and object-level correspondence task~\cite{yi2015otb}. Our experiments are implemented with MMAction2~\cite{2020mmaction2}. We will first introduce our training details and evaluation metrics, then we will perform extensive ablations on different elements for VFS and provides a better understanding on how VFS learns correspondence. Based on these observations, we finally report VFS surpasses all previous self-supervised learning approaches on both correspondence tasks. 

\subsection{Self-Supervised Pre-Training}

\noindent
\textbf{Architectures.} 
We use standard ResNet~\cite{he2016resnet} as the backbone network. We introduce the architecture for training without negative pairs (following~\cite{chen2020simsiam}) and the architecture for training with negative pairs (following~\cite{chen2020mocov2}) as below. 

\begin{itemize}[leftmargin=*]

\item \textit{Architecture without negative pairs.} 
The predictor encoder consists of a backbone network and a projector followed by a predictor. The target encoder is composed of the backbone and the projector. The parameters of the backbone and the projector are shared between the two encoders. The projector is a 3-layer MLP and the predictor is a 2-layer MLP as~\cite{chen2020simsiam}. All batch normalization layers in the backbone, the projector and the predictor are synchronized across devices (SyncBN) as in~\cite{chen2020simclr,grill2020byol,chen2020simsiam}.

\item \textit{Architecture with negative pairs.} 
The predictor encoder and target encoder share the same architecture, including a backbone followed by a 2-layer projector MLP. 
The parameters of the target encoder are updated with momentum $m=0.999$ with Eq.~\ref{eq:momentum}. There is no predictor head. Shuffle BN~\cite{chen2020mocov2} is used instead of SyncBN. We set the temperature $\tau=0.2$ and the negative bank size $K=65536$.

\end{itemize}

\noindent
\textbf{Pre-training.}
We adopt the  Kinetics\cite{kay2017kinetics} dataset for self-supervised training.
It consists of $\sim$240k training videos. 
The batch size is 256.
The learning rate is initialized to 0.05, and decays in the cosine schedule \cite{chen2020simclr, loshchilov2016sgdr,chen2020mocov2}. 
We use SGD optimizer with momentum 0.9 and weight decay 0.0001. 
We found that training for 100 epochs is sufficient for ResNet-18 models, 
and ResNet-50 models need 500 epochs to converge (roughly the same number of iterations as 100 epochs training~\cite{chen2020simsiam} on ImageNet).

\subsection{Evaluation}

We evaluate the pre-trained representation on both fine-grained and object-level correspondence downstream tasks.

\paragraph{Fine-grained Correspondence.}
To evaluate the quality of fine-grained correspondence, we follow the same testing protocol and downstream tasks in \cite{jabri2020walk,wang2019cycle}.
We re-implement the inference approach based on \cite{jabri2020walkcode}.
Without any fine-tuning, we directly use unsupervised pre-trained model as the feature extractor.
The fine-grained similarity is measured on the res$_4$ feature map, with its stride reduced to 1 during inference.
As in \cite{wang2019cycle,jabri2020walk, li2019uvc}, the recurrent inference strategy is applied: The first frame ground truth labels as well the prediction results in the latest 20 frames are propagated to the current frame. We evaluate the fine-grained correspondence over three downstream tasks and datasets: video object segmentation in DAVIS-2017~\cite{jordi2017davis}, human pose tracking in JHMDB~\cite{jhuang2013jhmdb} and human part tracking in VIP~\cite{zhou2018vip}. We perform most of our ablations with DAVIS-2017 and report the comparisons with state-of-the-art approaches in all datasets. 

\paragraph{Object-level Correspondence.}
We evaluate object-level correspondence with visual object tracking in the OTB-100~\cite{yi2015otb} dataset. We adopt the SiamFC~\cite{bertinetto2016siamfc} tracking algorithm with our representation from the res$_5$ block. We follow the same evaluation protocol and hyperparamters in VINCE~\cite{gordon2020vince} and SeCo~\cite{yao2021seco}: Given a pre-trained ResNet, the strides in res$_4$ and res$_5$ layers are removed, and the dilation rate of 3x3 convolution blocks in res$_4$ and res$_5$ layers are set to 2 and 4 respectively. This commonly used modification makes the res$_5$ block resolution compatible with the original setting in SiamFC. The network modification does not affect the pre-trained weights. We perform most of our analysis and ablation for VFS using the frozen representation after pre-training. We compare to the state-of-the-art results using fine-tuning in the end following~\cite{gordon2020vince,yao2021seco}.

\subsection{Results and Ablative Analysis}
\label{sec:ablation}

\begin{table}[t]
\centering
\small
\tablestyle{3pt}{1.1}
   \begin{tabular}{ccc|ccc|cc}
   \vspace{-3pt}
    &  &  & \multicolumn{3}{c|}{DAVIS}  & \multicolumn{2}{c}{OTB} \\
   {\tablestyle{0pt}{.9} \begin{tabular}{c} {different} \\ {frame} \end{tabular}}     
   & {\tablestyle{0pt}{.9} \begin{tabular}{c} {color} \\ {aug} \end{tabular}} 
   & {\tablestyle{0pt}{.9} \begin{tabular}{c} {spatial} \\ {aug} \end{tabular}} 
   & $\mathcal{J\&F}_m$ & $\mathcal{J}_m$ & $\mathcal{F}_m$ & Precision   & Success   \\
   \shline
         &        &        & 38.6 & 37.3 & 39.9 & 5.4  & 4.7  \\
         & \cmark &        & 50.9 & 49.3 & 52.6 & 0.4  & 0.3  \\
         &        & \cmark & 62.2 & 60.8 & 63.6 & 30.6 & 26.1 \\
         & \cmark & \cmark & 61.2 & 59.3 & 63.1 & 53.0 & 39.3 \\
    \hline
   \cmark &        &        & 63.4 & 61.1 & 65.7 & 37.4 & 28.8 \\
   \cmark & \cmark &        & 58.4 & 56.4 & 60.4 & 46.2 & 34.6 \\
   \cmark &        & \cmark & 65.0 & 62.6 & 67.4 & 48.1 & 37.9 \\
   \cmark & \cmark & \cmark & 61.9 & 59.5 & 64.3 & 57.3 & 43.0
   \end{tabular}
\vspace{.5em}
\caption{
  \textbf{Ablation on different augmentations}. 
  Color augmentations are random color jitter, grayscale conversion, gaussian blur. 
  Spatial augmentations are random resized crop and horizontal flip. 
  ``different frame'' indicates whether the inputs for predictor and target encoder are different.
  \label{tab:aug}
\vspace{-.8em}
}
\end{table}

\begin{table}[t]
\centering
\small
\tablestyle{8pt}{1.1}
   \begin{tabular}{c|ccc|cc}
   \vspace{-3pt}
   & \multicolumn{3}{c|}{DAVIS}  & \multicolumn{2}{c}{OTB} \\
   {\tablestyle{0pt}{.9} \begin{tabular}{c} {frame} \\ {interval} \end{tabular}} 
   & $\mathcal{J\&F}_m$ & $\mathcal{J}_m$ & $\mathcal{F}_m$ & Precision   & Success   \\
   \shline
   0  & 62.2 & 60.8 & 63.6 & 30.6 & 26.1 \\
   2  & 63.5 & 61.7 & 65.4 & 38.1 & 31.2 \\
   4  & 62.9 & 61.4 & 64.3 & 35.5 & 29.5 \\
   8  & 63.5 & 61.7 & 65.3 & 38.5 & 31.8 \\
   16 & 63.9 & 61.9 & 65.8 & 44.1 & 33.9 \\
   32 & 64.5 & 62.4 & 66.7 & 46.9 & 36.1 \\
   D  & 65.0 & 62.6 & 67.4 & 48.1 & 37.9
  \end{tabular}
\vspace{.5em}
\caption{
  \textbf{Ablation on frame interval}. 
  0 means sample two identical frames. D stands for the \textit{distant sampling}. 
  Others use \textit{continuous sampling} with fixed frame sampling interval.
  \label{tab:gap}
\vspace{-1.5em}
}
\end{table}

We first perform analysis on different elements of VFS. We use ResNet-18 as the default backbone. The experiments are under the \textit{without negative pairs} setting unless specified otherwise. 

\paragraph{Augmentation.}

Out first discovery is that color augmentation plays an important role in VFS and affect fine-grained and object-level correspondence in an opposite way. We report our results on augmentations in Table~\ref{tab:aug}. With color augmentation, the performance on DAVIS decrease by $> 3\%$ and the OTB precision improves over $10\%$. We conjecture that while color augmentation helps learning invariance to object appearance changes, the per-pixel distortion confuses the lower-level convolution features to find fine-grained correspondence. From Table~\ref{tab:aug}, we also observe sampling different frames in a video (with distant sampling) indeed help learning both object-level and fine-grained correspondence, and adding spatial augmentations (random crop and flip) can further improve the results. In the following experiments, we adopt different frame inputs and spatial augmentation by default. We will report results with and without color augmentations in the different ablations.

\paragraph{Temporal Sampling.} Does sampling with larger frame interval improve correspondence learning? To answer this question, we study the temporal sampling strategy and report our results in Table~\ref{tab:gap}. We perform the study without using the color augmentation. Recall we have two temporal sampling strategies in VFS (Section~\ref{sec:vfs}). With \textit{continuous sampling}, we elaborate the frame interval $\delta$ from 0 to 32. We observe as $\delta$ increases, both fine-grained and object-level correspondence improves consistently. With \textit{distant sampling} using $n=2$ frame inputs (labeled as ``D'' in Table~\ref{tab:gap}), we achieve the best results. We conjecture increasing the frame intervals offers better augmentation effect which leads to better correspondence. We will adopt the distant sampling in the following subsections.

\begin{table}[t]
\centering
\small
\tablestyle{6pt}{1.1}
   \begin{tabular}{cc|ccc|cc}
   \vspace{-3pt}
  &  & \multicolumn{3}{c|}{DAVIS}  & \multicolumn{2}{c}{OTB} \\
  {\tablestyle{0pt}{.9} \begin{tabular}{c} {color} \\ {aug} \end{tabular}} 
  & {\tablestyle{0pt}{.9} \begin{tabular}{c} {frame} \\ {num} \end{tabular}} 
  & $\mathcal{J\&F}_m$ & $\mathcal{J}_m$ & $\mathcal{F}_m$ & Precision   & Success   \\
  \shline
         & n=2 & 65.0 & 62.6 & 67.4 & 48.1 & 37.9 \\
         & n=4 & 65.8 & 63.2 & 68.4 & 51.5 & 38.4 \\
         & n=8 & 66.7 & 64.0 & 69.4 & 53.0 & 39.6 \\
  \hline
  \cmark & n=2 & 61.9 & 59.5 & 64.3 & 57.3 & 43.0 \\
  \cmark & n=4 & 62.9 & 60.5 & 65.3 & 58.4 & 43.8 \\
  \cmark & n=8 & 62.5 & 59.8 & 65.1 & 59.0 & 43.8
  \end{tabular}
\vspace{.5em}
\caption{
  \textbf{Ablation on multiple frames}. 
  ``frame num'' is the number of sampled frame.
  \label{tab:multi}
\vspace{-.8em}
}
\end{table}

\begin{table}[t]
\centering
\small
\tablestyle{3pt}{1.1}
   \begin{tabular}{cc|ccc|cc|c}
   \vspace{-3pt}
  &  & \multicolumn{3}{c|}{DAVIS}  & \multicolumn{2}{c}{OTB} & ImageNet\\
  {\tablestyle{0pt}{.9} \begin{tabular}{c} {negative} \\ {pairs} \end{tabular}} 
  & {\tablestyle{0pt}{.9} \begin{tabular}{c} {color} \\ {aug} \end{tabular}} 
  & $\mathcal{J\&F}_m$ & $\mathcal{J}_m$ & $\mathcal{F}_m$ & Precision   & Success   & Acc@1\\
  \shline
       &        & 65.0 & 62.6 & 67.4 & 48.1 & 37.9 & 22.0 \\
\cmark &        & 64.7 & 62.2 & 67.3 & 39.0 & 31.7 & 24.2 \\
  \hline
       & \cmark & 61.9 & 59.5 & 64.3 & 57.3 & 43.0 & 31.8 \\
\cmark & \cmark & 61.5 & 59.3 & 63.7 & 53.7 & 40.8 & 33.8
  \end{tabular}
\vspace{.5em}
\caption{
  \textbf{Ablation on negative pairs}. For \textit{with negative pairs} setting, the learning objective is Eq~\ref{eq:infonce_vid}.
  \label{tab:negative}
\vspace{-.8em}
}
\end{table}

\begin{table}[t]
\centering
\small
\tablestyle{3pt}{1.1}
   \begin{tabular}{ccc|ccc|cc}
   \vspace{-3pt}
  &  &  & \multicolumn{3}{c|}{DAVIS}  & \multicolumn{2}{c}{OTB} \\
  {\tablestyle{0pt}{.9} \begin{tabular}{c} {negative} \\ {pairs} \end{tabular}} 
  & {\tablestyle{0pt}{.9} \begin{tabular}{c} {color} \\ {aug} \end{tabular}} 
  & Backbone
  & $\mathcal{J\&F}_m$ & $\mathcal{J}_m$ & $\mathcal{F}_m$ & Precision   & Success   \\
  \shline
         &        & ResNet-18 & 65.0 & 62.6 & 67.4 & 48.1 & 37.9 \\
         &        & ResNet-50 & 68.9 & 66.5 & 71.3 & 47.4 & 34.6 \\
  \cmark &        & ResNet-18 & 64.7 & 62.2 & 67.3 & 39.0 & 31.7 \\
  \cmark &        & ResNet-50 & 68.3 & 65.8 & 70.8 & 46.4 & 34.4 \\
  \hline
         & \cmark & ResNet-18 & 61.9 & 59.5 & 64.3 & 57.3 & 43.0 \\
         & \cmark & ResNet-50 & 67.1 & 64.6 & 69.6 & 59.5 & 43.4 \\
  \cmark & \cmark & ResNet-18 & 61.5 & 59.3 & 63.7 & 53.7 & 40.8 \\
  \cmark & \cmark & ResNet-50 & 67.2 & 64.7 & 69.7 & 56.5 & 40.7
  \end{tabular}
\vspace{.5em}
\caption{
  \textbf{Ablation on deeper models}. ResNet-18 and ResNet-50 are compared.
  \label{tab:deeper}
\vspace{-1.5em}
}
\end{table}

\paragraph{Multiple Frames.} We investigate the effect of training with multiple pairs of frames from a video. With distant sampling, we ablate the number of segments $n=2,4,8$, which indicates using $1,2,4$ pairs of frames from one video during training. We report our results in Table~\ref{tab:multi}. With or without using color augmentation, we observe training with more pairs of frames generally improves the correspondence representation. Note the first row in Table~\ref{tab:multi} corresponds to the last row in Table~\ref{tab:gap}. 

\paragraph{Negative Pairs.} So far we conduct our experiments under the \textit{without negative pairs} setting. But will adding the negative pairs (i.e., contrastive learning) help correspondence learning? To our surprise, adding negative pairs in training hurts learning both fine-grained and object-level correspondence. We report the comparisons in Table~\ref{tab:negative}. While the fine-grained correspondence result on DAVIS drops slightly with negative pairs, the performance of object tracking in OTB degenerates significantly. When training without color augmentations, the object tracking precision drops $\sim9\%$. 

What is the reason causing this performance drop when training with negatives? Our hypothesis is that training with negative pairs may sacrifice the performance on modeling intra-instance invariance for learning better features for cross instance discrimination. To prove this hypothesis, we perform linear classification on top of the frozen features on the  ImageNet-1k dataset~\cite{deng2009imagenet} and report the results on the right column of Table~\ref{tab:negative}. We observe that the model trained with negatives indeed leads to better semantic classification with around $2\%$ improvement, which supports our hypothesis. Note that these results are not directly comparable to state-of-the-art results on ImageNet-1k~\cite{chen2020simclr,chen2020mocov2,chen2020simsiam}, since we train our model (ResNet-18 backbone) with video frames (with large domain gap) and optimize for correspondence learning instead of semantic classification. With these observations, we conjecture training without negatives can learn better correspondence representation, which has not been shown in previous literature to our knowledge.

\begin{figure}[t]
\vspace{-1em}
\centering
\includegraphics[width=.9\linewidth]{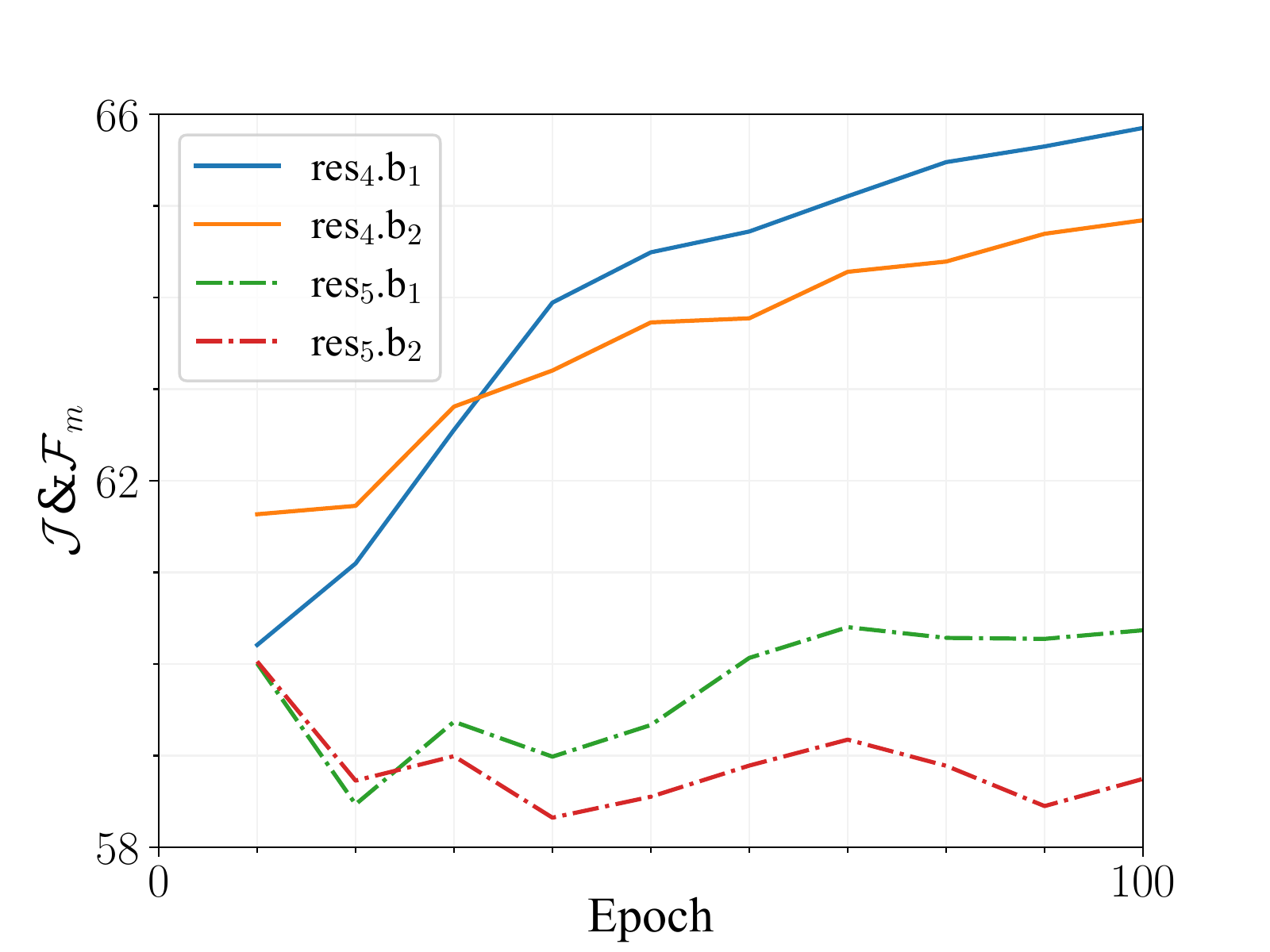}
\caption{
$\mathcal{J\&F}_m$ on DAVIS at different epochs with \textbf{ResNet-18}. 
\label{fig:layers}
}
\vspace{-1.5em}
\end{figure}

\begin{figure}[t]
\vspace{-.5em}
\centering
\includegraphics[width=.9\linewidth]{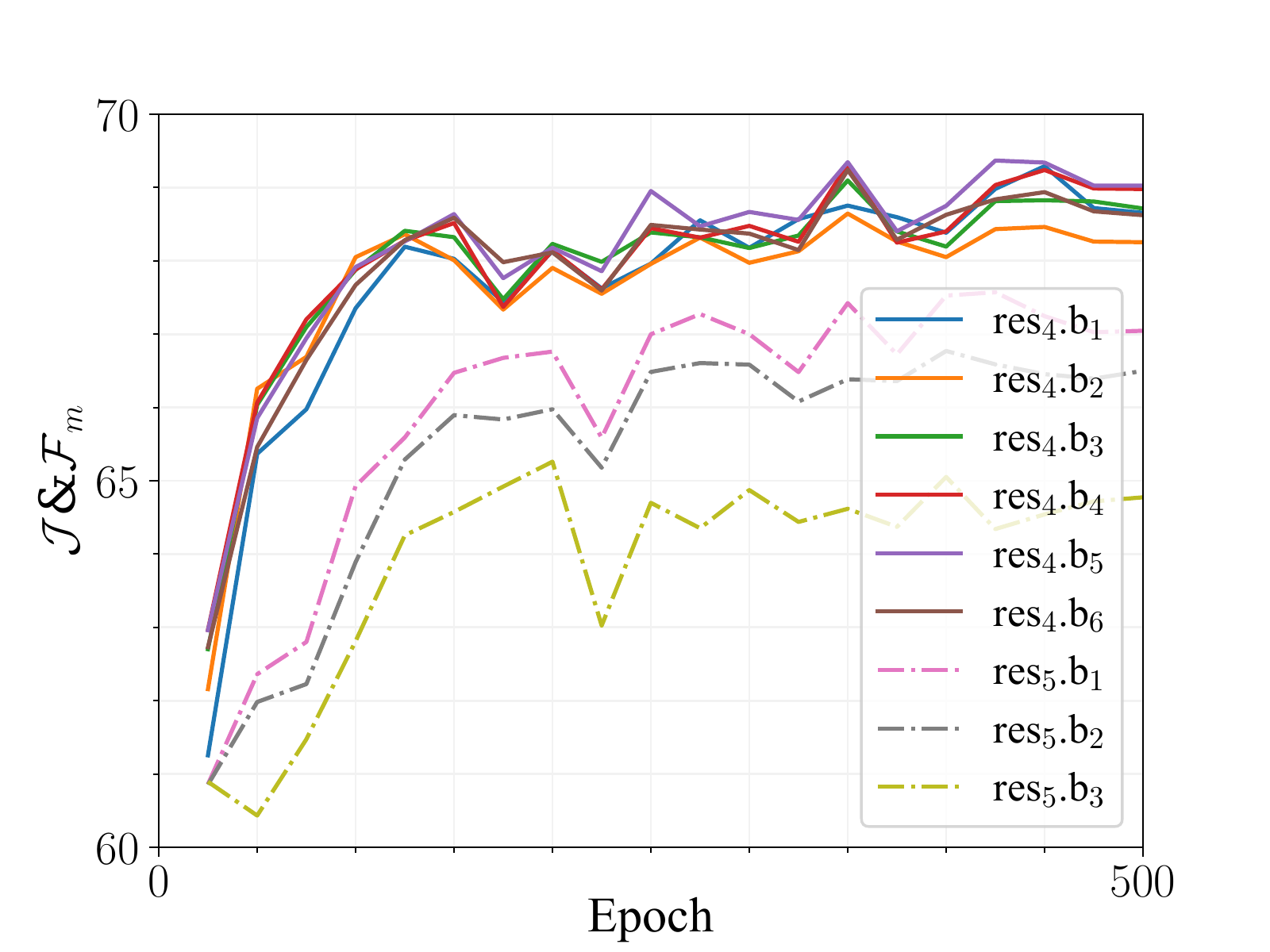}
\caption{
$\mathcal{J\&F}_m$ on DAVIS at different epochs with \textbf{ResNet-50}. 
\label{fig:layers_r50}
}
\vspace{-2.5em}
\end{figure}

\paragraph{Deeper Models.}
We so far mainly perform analysis with ResNet-18.  We observe the performance does not change or even degrades in previous self-supervised learning approaches~\cite{jabri2020walk,li2019uvc, lai2020mast, wang2019cycle} with ResNet-50. So can VFS be scaled up to deeper models? To answer this, we train  VFS with ResNet-50 backbone and report the results in Table~\ref{tab:deeper}. We observe deeper networks can significantly improve both fine-grained and object-level correspondence with VFS.  
Under without negative pairs setting, ResNet-50 backbone improves DAVIS $\mathcal{J\&F}_m$ by $3.9\%$, OTB precision by $2.2\%$ over ResNet-18.
More interestingly, while the ResNet-18 performance on DAVIS decrease by $3.1\%$ with color augmentation, the gap has been closed to $1.8\%$ under the ResNet-50 setting. We conjecture stacking more convolution blocks helps adapt to appearance distortion. 
More ResNet-50 comparison could be found in Table~\ref{tab:sota-pixel} and \ref{tab:sota-object}. 

\paragraph{Different Blocks of Layers.} We plot the VFS model performance on DAVIS throughout different epochs of similarity learning with ResNet-18 (Figure~\ref{fig:layers}) and ResNet-50 (Figure~\ref{fig:layers_r50}). We investigate how each feature block in res$_4$ and res$_5$ layers performs. For example, res$_4$,b$_1$ indicates the feature from the first block in res$_4$. In Figure~\ref{fig:layers}, all blocks begin with $\sim 60\%$ on DAVIS in early stages of training. However, the gap between res$_4$ and res$_5$ becomes larger as the model is trained longer. The blocks in res$_5$ ends up with similarly scores as beginning, while the results of res$_4$ continuously improve. This supports our previous results that res$_4$ learns fine-grained correspondence over time and res$_5$ focuses more on object-level features. Similarly for ResNet-50 in Figure~\ref{fig:layers_r50}, res$_4$ also outperforms res$_5$ by a noticeable margin for finding fine-grained correspondence.

\subsection{Comparison with State-Of-The-Art}

\begin{table*}[!htb]
\centering
\small
\tablestyle{5pt}{1.1}
   \begin{tabular}{llcc|ccc|c|cc}
   \vspace{-3pt}
  &  &  &  & \multicolumn{3}{c|}{DAVIS}  & \multicolumn{1}{c|}{VIP} & \multicolumn{2}{c}{JHMDB}\\
  Method
  & Backbone
  & Stride
  & Dataset
  & $\mathcal{J\&F}_m$ & $\mathcal{J}_m$ & $\mathcal{F}_m$ & mIoU   & PCK@0.1 & PCK@0.2   \\
  \shline
  Supervised~\cite{he2016resnet} & ResNet-18            & 32 & ImageNet & 62.9 & 60.6 & 65.2 & 31.9 & 53.8 & 74.6 \\
  SimSiam~\cite{chen2020simsiam} & ResNet-18            & 32 & ImageNet & 62.0 & 60.0 & 64.0 & 30.3 & 55.2 & 74.0 \\
  MoCo~\cite{he2020moco}         & ResNet-18            & 32 & ImageNet & 60.8 & 58.6 & 63.1 & 29.2 & 55.0 & 72.7 \\
  VINCE~\cite{gordon2020vince}   & ResNet-18            & 32 & Kinetics & 60.4 & 57.9 & 62.8 & 30.5 & 55.6 & 73.4 \\
  \hline
  CorrFlow~\cite{lai2019corrflow}$^*$ & ResNet-18$^\ddagger$ & 4 & OxUvA    & 50.3 & 48.4 & 52.2 & -    & 58.5 & 78.8 \\
  MAST~\cite{lai2020mast}$^*$    & ResNet-18$^\ddagger$ & 4  & OxUvA    & 63.7 & 61.2 & 66.3 & -    & -    & -    \\
  MAST~\cite{lai2020mast}$^*$    & ResNet-18$^\ddagger$ & 4  & YT-VOS   & 65.5 & 63.3 & 67.6 & -    & -    & -    \\
  \hline
  Vid. Color.~\cite{carl2018color}    & ResNet-18$^\dagger$   & 8 & Kinetics & 34.0 & 34.6 & 32.7 & -    & 45.2 & 69.6 \\
  TimeCycle~\cite{wang2019cycle}      & ResNet-18$^\dagger$   & 8 & VLOG     & 39.2 & 40.1 & 38.3 & 28.9 & 57.3 & 78.1 \\
  UVC~\cite{li2019uvc}           & ResNet-18$^\dagger$   & 8  & Kinetics & 57.8 & 56.3 & 59.2 & 34.1 & 58.6 & 79.6 \\
  UVC+track~\cite{li2019uvc}$^*$ & ResNet-18$^\dagger$   & 8  & Kinetics & 59.5 & 57.7 & 61.3 & -    & -    & -    \\
  CRW~\cite{jabri2020walk}            & ResNet-18$^\dagger$   & 8 & Kinetics & \textbf{67.6} & \textbf{64.8} & \textbf{70.2} & 38.6 & 59.3 & \textbf{80.3} \\
  \textbf{VFS}                   & ResNet-18            & 32 & Kinetics & 66.7 & 64.0 & 69.4 & \textbf{39.9} & \textbf{60.5} & 79.5 \\
  \color{Gray}{VFS (best block)}     & \color{Gray}{ResNet-18}            & \color{Gray}{32} & \color{Gray}{Kinetics} & \color{Gray}{67.9} & \color{Gray}{65.0} & \color{Gray}{70.8} & \color{Gray}{-}    & \color{Gray}{-}    & \color{Gray}{-}    \\
  \hhline{==========}
  Supervised~\cite{he2016resnet} & ResNet-50            & 32 & ImageNet & 66.0 & 63.7 & 68.4 & 39.5 & 59.2 & 78.3 \\
  SimSiam~\cite{chen2020simsiam} & ResNet-50            & 32 & ImageNet & 66.3 & 64.5 & 68.2 & 35.0 & 58.4 & 77.5 \\
  MoCo~\cite{he2020moco}         & ResNet-50            & 32 & ImageNet & 65.4 & 63.2 & 67.6 & 36.1 & 60.4 & 79.3 \\
  VINCE~\cite{gordon2020vince}   & ResNet-50            & 32 & Kinetics & 65.6 & 63.4 & 67.8 & 36.0 & 58.2 & 76.3 \\
  RegionTracker~\cite{purushwalkam2020demystifying} & ResNet-50 & 32 & TrackingNet & 63.4 & 61.5 & 65.4 & 33.9 & 57.5 & 74.6 \\
  \hline
  TimeCycle~\cite{wang2019cycle}      & ResNet-50$^\dagger$  & 8 & VLOG     & 40.7 & 41.9 & 39.4 & 28.9 & 57.7 & 78.5 \\
  UVC~\cite{li2019uvc}           & ResNet-50$^\dagger$  & 8  & Kinetics & 56.3 & 54.5 & 58.1 & 34.2 & 56.0 & 76.6 \\
  \textbf{VFS}                   & ResNet-50            & 32 & Kinetics & \textbf{68.9} & \textbf{66.5} & \textbf{71.3} & \textbf{43.2} & \textbf{60.9} & \textbf{80.7} \\
  \color{Gray}{VFS (best block) }                  & \color{Gray}{ResNet-50}            & \color{Gray}{32} & \color{Gray}{Kinetics} & \color{Gray}{69.4} & \color{Gray}{66.7} & \color{Gray}{72.0} & \color{Gray}{-} & \color{Gray}{-} & \color{Gray}{-}
  \end{tabular}
\vspace{0.5em}
\caption{
  \textbf{Comparison with state-of-the-art on fine-grained correspondence}. $*$ indicates localization is involved during label propagation. 
  $\dagger$ denotes strides of last two layers are removed. 
  $\ddagger$ denotes max pooling of stem layer is also removed. 
  Stride is the output stride (downsample ratio) of ResNet during training. 
  \label{tab:sota-pixel}
\vspace{-1.5em}
}
\end{table*}

\begin{table}[t]
\centering
\tablestyle{5pt}{1.1}
   \begin{tabular}{llc|cc}
   \vspace{-3pt}
  &  &  & \multicolumn{2}{c}{OTB}\\
  Method
  & Backbone
  & Dataset
  & Precision & Success   \\
  \shline
  Supervised~\cite{he2016resnet} & ResNet-18           & ImageNet & 61.4 & 43.0 \\
  SimSiam~\cite{chen2020simsiam} & ResNet-18           & ImageNet & 58.8 & 42.9 \\
  MoCo~\cite{he2020moco}         & ResNet-18           & ImageNet & 62.0 & 47.0 \\
  VINCE~\cite{gordon2020vince}   & ResNet-18           & Kinetics & 62.9 & 46.5 \\
  CRW~\cite{jabri2020walk}       & ResNet-18$^\dagger$ & Kinetics & 52.6 & 40.1 \\
  \textbf{VFS}                   & ResNet-18           & Kinetics & \textbf{68.9} & \textbf{52.2} \\
  \hhline{=====}
  Supervised~\cite{he2016resnet} & ResNet-50           & ImageNet & 65.8 & 45.5 \\
  SimSiam~\cite{chen2020simsiam} & ResNet-50           & ImageNet & 61.0 & 43.2 \\
  MoCo~\cite{he2020moco}         & ResNet-50           & ImageNet & 63.7 & 46.5 \\
  VINCE~\cite{gordon2020vince}   & ResNet-50           & Kinetics & 66.0 & 47.6 \\
  RegionTracker~\cite{purushwalkam2020demystifying} & ResNet-50 & TrackingNet & 57.4 & 43.4 \\
  SeCo~\cite{yao2021seco}        & ResNet-50           & Kinetics & 71.9 & 51.8 \\
  \textbf{VFS}                   & ResNet-50           & Kinetics & \textbf{73.9} & \textbf{52.5}
  \end{tabular}
\vspace{0.5em}
\caption{
  \textbf{Comparison with state-of-the-art on object-level correspondence}. 
  $\dagger$ denotes strides of last two layers are removed. 
  \label{tab:sota-object}
\vspace{-2.5em}
}
\end{table}

We compare fine-grained correspondence results of VFS against previous self-supervised methods in Table~\ref{tab:sota-pixel}. The results are all reported with the last block in res$_4$ across all methods. Our method achieves state-of-the-art performance using ResNet-50. With ResNet-50, we observes UVC~\cite{li2019uvc} does not benefit from using a deeper networks. But for CRW~\cite{jabri2020walk} we couldn't make it work on large models. Learning with  VFS, the deeper network with ResNet-50 improves $2.2\%$ on DAVIS and $3.3\%$ on VIP over ResNet-18, which is significant. We observe consistent results across the JHMDB~\cite{jhuang2013jhmdb} human pose and VIP~\cite{zhou2018vip} human part tracking tasks. With ResNet-18, VFS achieves comparable performance with CRW~\cite{jabri2020walk}. As shown in Figure~\ref{fig:layers} and \ref{fig:layers_r50}, the last block in res$_4$ may not achieve the optimal performance, thus we also report the result of the best block with gray color for reference.

The comparison on learning object-level correspondence is reported in Table~\ref{tab:sota-object}.  
We report the fine-tuning performance of our best setting, namely distant sampling with $n=8$, with color augmentation and without negative pairs. 
VFS with ResNet-50 backbone brings $5\%$ precision gain over ResNet-18, yields $73.9\%$ precision and $52.5\%$ success score. 
Our simple VFS surpasses ImageNet supervised pre-training as well as previous self-supervised state-of-the-art  SeCo~\cite{yao2021seco}, which involves joint training of 3 pretext tasks.

\section{Discussion and Conclusion}

We propose a simple yet effective approach for self-supervised correspondence learning.
We demonstrate that both fine-grained and object-level correspondence can emerge in different layers of the ConvNets with video frame-level similarity learning. In addition to the state-of-the-art performance, we provide the following insights. 

\textit{\textbf{Is designing a tracking-based pretext task a necessity for self-supervised correspondence learning?}} It might not be necessary. While tracking-based pretext tasks still have potentials, it is limited by small backbone models and is now surpassed by our simple frame-level similarity learning. To make the tracking-based pretext tasks useful, we need to first make its learning scalable and generalizable in model size and network architectures.

\textit{\textbf{Does color augmentation improve the correspondence?}} Yes and no. 
We show that color augmentation is beneficial for correspondence in object-level but jeopardizes the fine-grained correspondence. While color augmentation brings object appearance invariance, it also confuses the lower-layer convolution features.

\textit{\textbf{How to sample video frames?}} Sample multiple frames and sample with a large gap. 
The large temporal gap provides more aggressive temporal transform, which boosts correspondence significantly. 
Comparing multiple pairs of frame further improves the results.

\textit{\textbf{Is negative pairs helpful?}} No. 
We observe inferior performance when training with negative samples, specifically for object-level correspondence.
We also shed light on the reason why \textit{without negative pairs} is more helpful, which has not been studied before.

In conclusion, we hope VFS can serve as a strong baseline for self-supervised correspondence learning.  We provide a new perspective on studying similarity learning for visual representations beyond recognition tasks. 

\vspace{1em}
{\footnotesize \textbf{Acknowledgements.}~This work was supported, in part, by grants from DARPA LwLL, NSF 1730158 CI-New: Cognitive Hardware and Software Ecosystem Community Infrastructure (CHASE-CI), NSF ACI-1541349 CC*DNI Pacific Research Platform, and gifts from Qualcomm and TuSimple.}


\renewcommand\thefigure{\thesection.\arabic{figure}}
\renewcommand\thetable{\thesection.\arabic{table}}
\setcounter{figure}{0} 
\setcounter{table}{0} 

{\small
\bibliographystyle{ieee_fullname}
\bibliography{rethink_bib}
}

\newpage
\appendix

\section{Implementation Details}

\paragraph{Fine-grained Correspondence}
We apply recurrent inference strategy for fine-grained correspondence. 
To be more specific, we calculate the similarity between the current frame with the first frame ground truth labels as well the prediction results in the preceding $m$ frames. 
Then the labels of top-k most similarly pixels are selected and propagated to the current frame. 
We only compute the similarity between features that are at most $r$ pixels away from each other, i.e. \textit{local} attention. 
The detailed hyperparamter setting for each dataset are listed in Table~\ref{tab:hyper}

\begin{table}[!h]
\centering
\small
\tablestyle{3pt}{1.1}
\begin{tabular}{l|ccc}
                    & DAVIS & VIP & JHMDB \\
\shline
top-k              & 10    & 10  & 10    \\
preceding frame $m$    & 20    & 8   & 4     \\
propagation radius $r$ & 12    & 20  & 20   
\end{tabular}
\vspace{0.5em}
\caption{Fine-grained Correspondence Inference Hyperparameter.
\label{tab:hyper}}
\vspace{-0.8em}
\end{table}

\paragraph{Object-level Correspondence}
For the fair comparison, we use fine-tuning setting when comparing with previous approaches~\cite{gordon2020vince,yao2021seco}.
Specifically, an additional $1\times 1$ convolution is placed on top of the backbone to transform the frozen representation.
Note that only this $1\times 1$ convolution is learnable during fine-tuning. 
So such protocol could be considered as the linear evaluation.
We fine-tune the the $1\times 1$ convolution layer on the GOT-10K~\cite{huang2019got} dataset, which consists of $\sim$ 10,000 video clips and 1.4 million frames.
Adam optimizer is adopted during fine-tuning. 
The learning rate is initialized to 0.001 and decays by 0.9 every epoch.
There is no weight decay.
The network is fine-tuned for 50 epochs. 
The batch size is 8 for all experiments.
The inference hyperparamters are the same for without fine-tuning and with fine-tuning setting.

\section{Visualization}

\begin{figure*}[t]
\centering
\includegraphics[width=.9\linewidth]{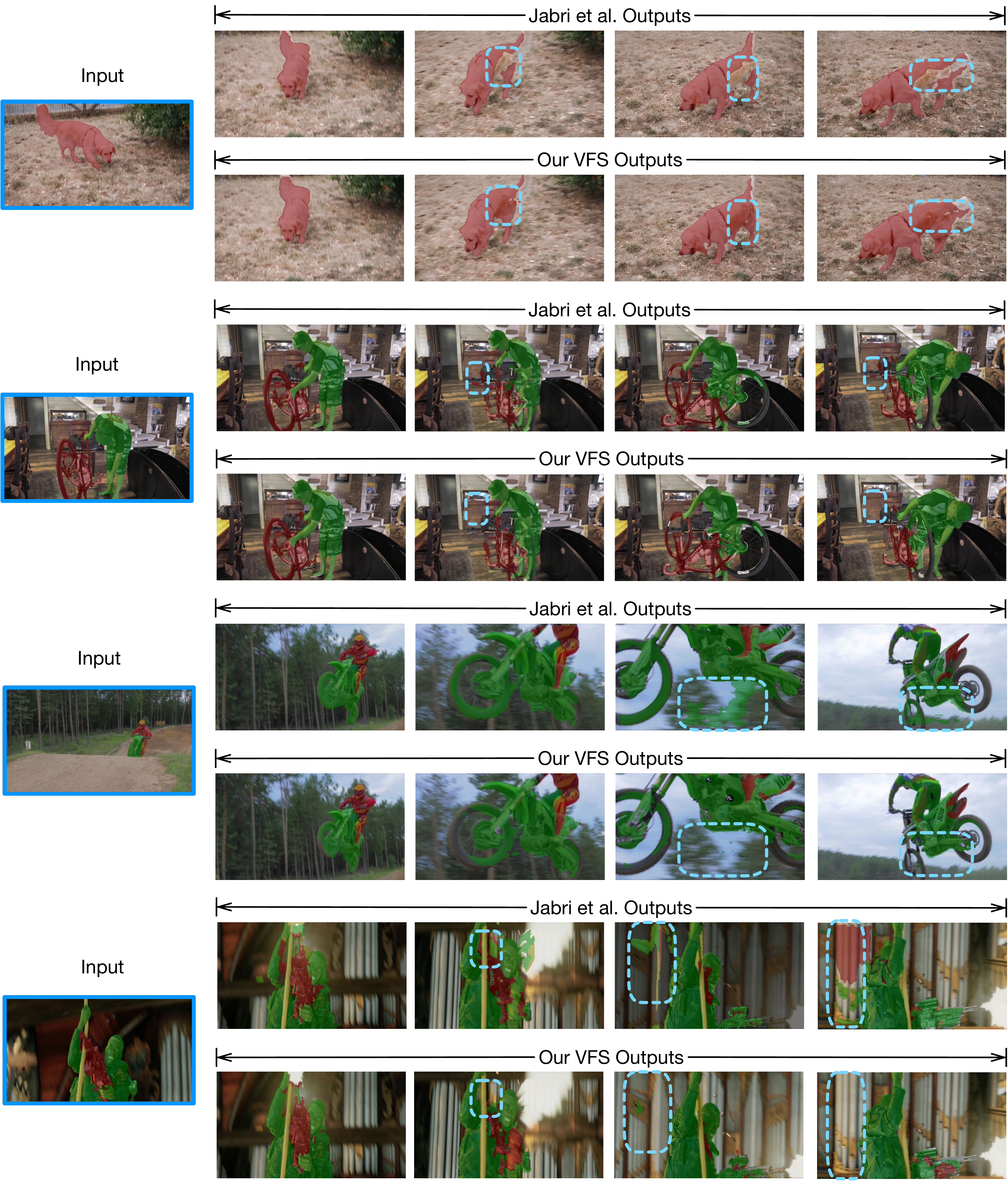}
\vspace{.5em}
\caption{
\textbf{Compare Fine-grained Correspondence on DAVIS}. 
Comparing with previous state-of-the-art Jabri et al.~\cite{jabri2020walk}, our VFS could generate results of higher quality and with less false positives.
Blue dash areas indicate failure cases in \cite{jabri2020walk}, where our approach could output plausible results.
More comparison are provided in the \href{https://jerryxu.net/VFS/}{project page}.
\label{fig:vis_comp}
}
\end{figure*}
\begin{figure*}[t]
\centering
\includegraphics[width=.99\linewidth]{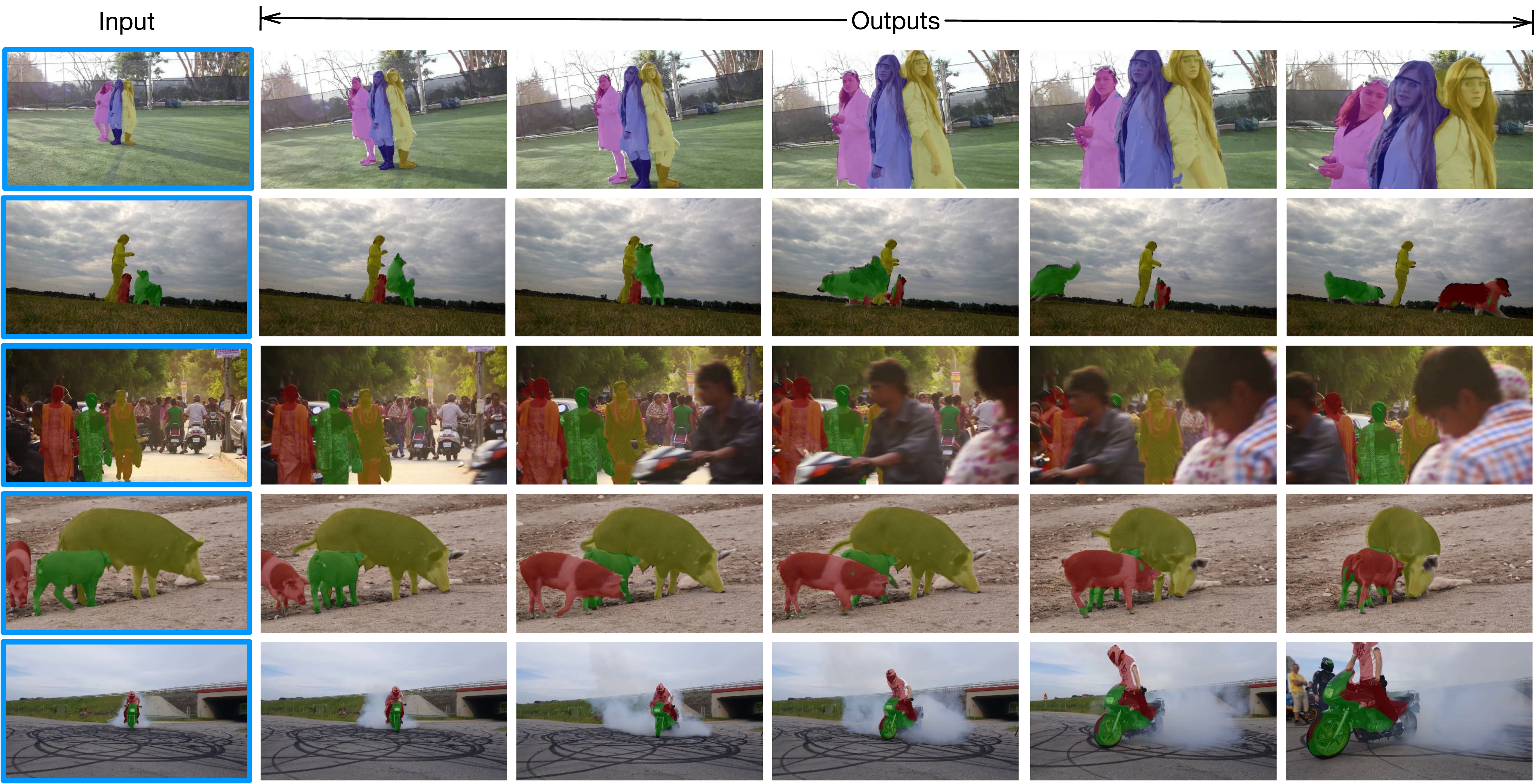}
\vspace{.2em}
\caption{
Qualitative Results for video object segmentation on DAVIS-2017~\cite{jordi2017davis}. 
\label{fig:vis_davis}
\vspace{-1.5em}
}
\end{figure*}
\begin{figure*}[t]
\centering
\includegraphics[width=.99\linewidth]{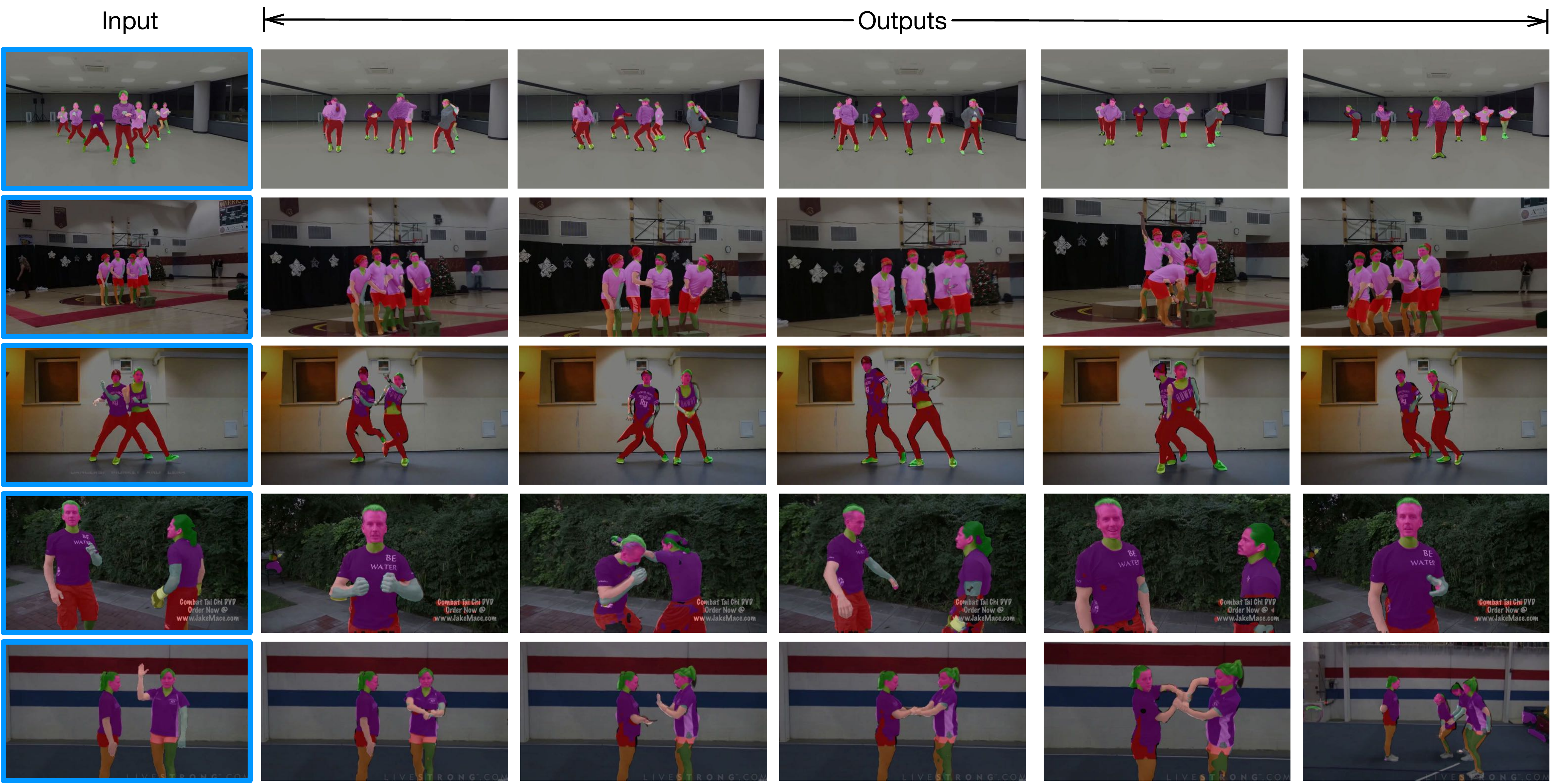}
\vspace{.2em}
\caption{
Qualitative Results for human part tracking on VIP~\cite{zhou2018vip}. 
\label{fig:vis_vip}
\vspace{-1.5em}
}
\end{figure*}
\begin{figure*}[t]
\centering
\includegraphics[width=.88\linewidth]{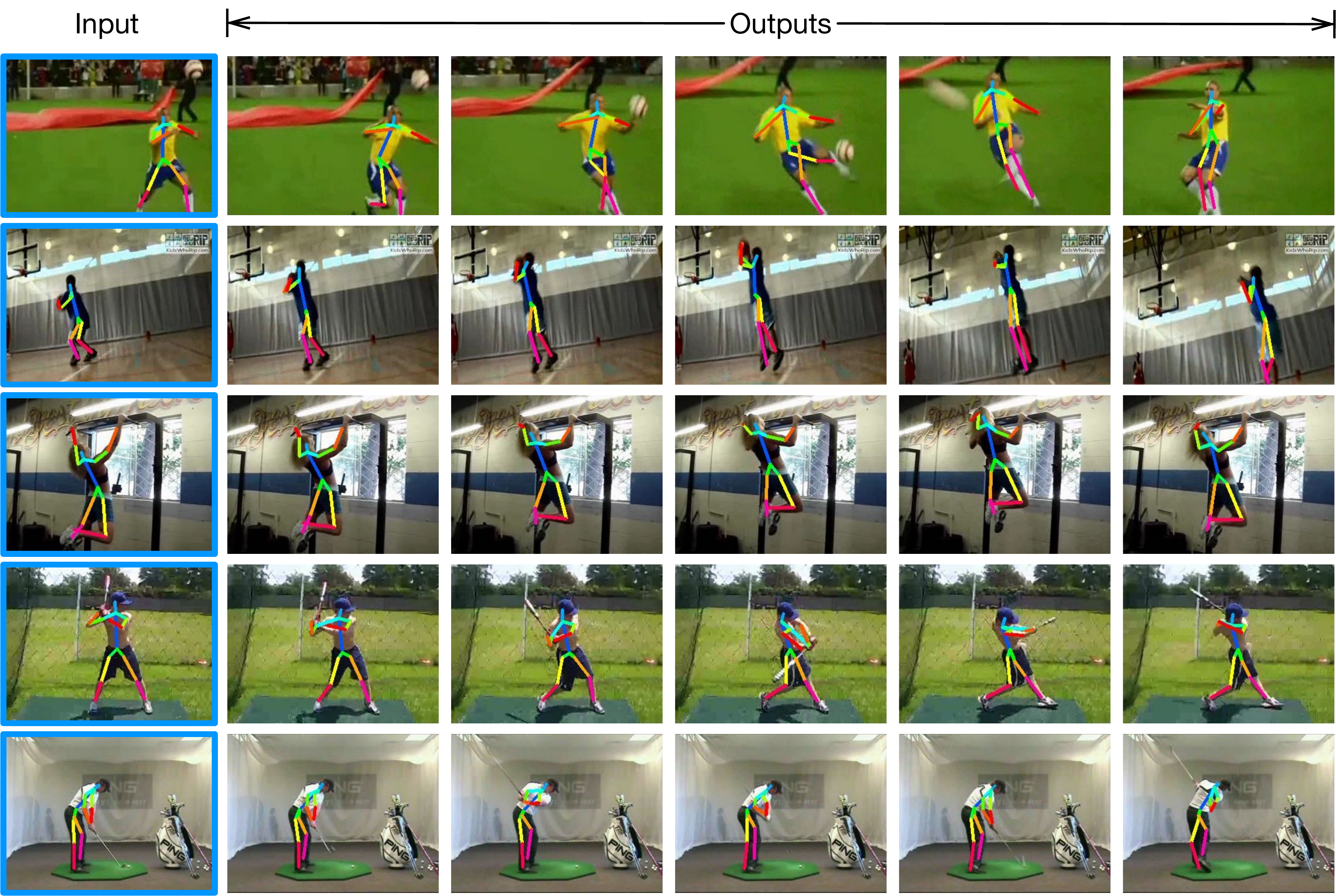}
\vspace{.2em}
\caption{
Qualitative Results for human pose tracking on JHMDB~\cite{jhuang2013jhmdb}. 
\label{fig:vis_jhmdb}
\vspace{-1.5em}
}
\end{figure*}

Without fine-tuning on any additional dataset, the fine-grained correspondence are directly evaluated on the res$_4$ features of pre-trained ResNet.
We visualize our correspondence on 3 downstream tasks and datasets in Figure~\ref{fig:vis_davis},\ref{fig:vis_jhmdb},\ref{fig:vis_vip}, i.e. video object segmentation on DAVIS-2017~\cite{jordi2017davis}, human pose tracking on JHMDB~\cite{jhuang2013jhmdb}, and human part tracking on VIP~\cite{zhou2018vip}.
For DAVIS and VIP, there are usually more than one instances/parts. 
Our approach could output tight boundaries around the multiple target areas.
For example, in the last row of Figure~\ref{fig:vis_vip}, the human parts could still be segmented when more people appears in the video.
In human pose tracking, even though each joint is propagated individually, we could still estimate the pose accurately.
We also compare our VFS with state-of-the-art method \cite{jabri2020walk} in Figure~\ref{fig:vis_comp}. 
As last three rows illustrated, our VFS has less false positive object segmentation than \cite{jabri2020walk}.
It indicates that our VFS is more robust to distinguish similar pixels.
Note that the inference hyperparamters for both methods are the same, the only difference is the pre-trained representation weight.

\begin{figure*}[!h]
\centering
\includegraphics[width=.88\linewidth]{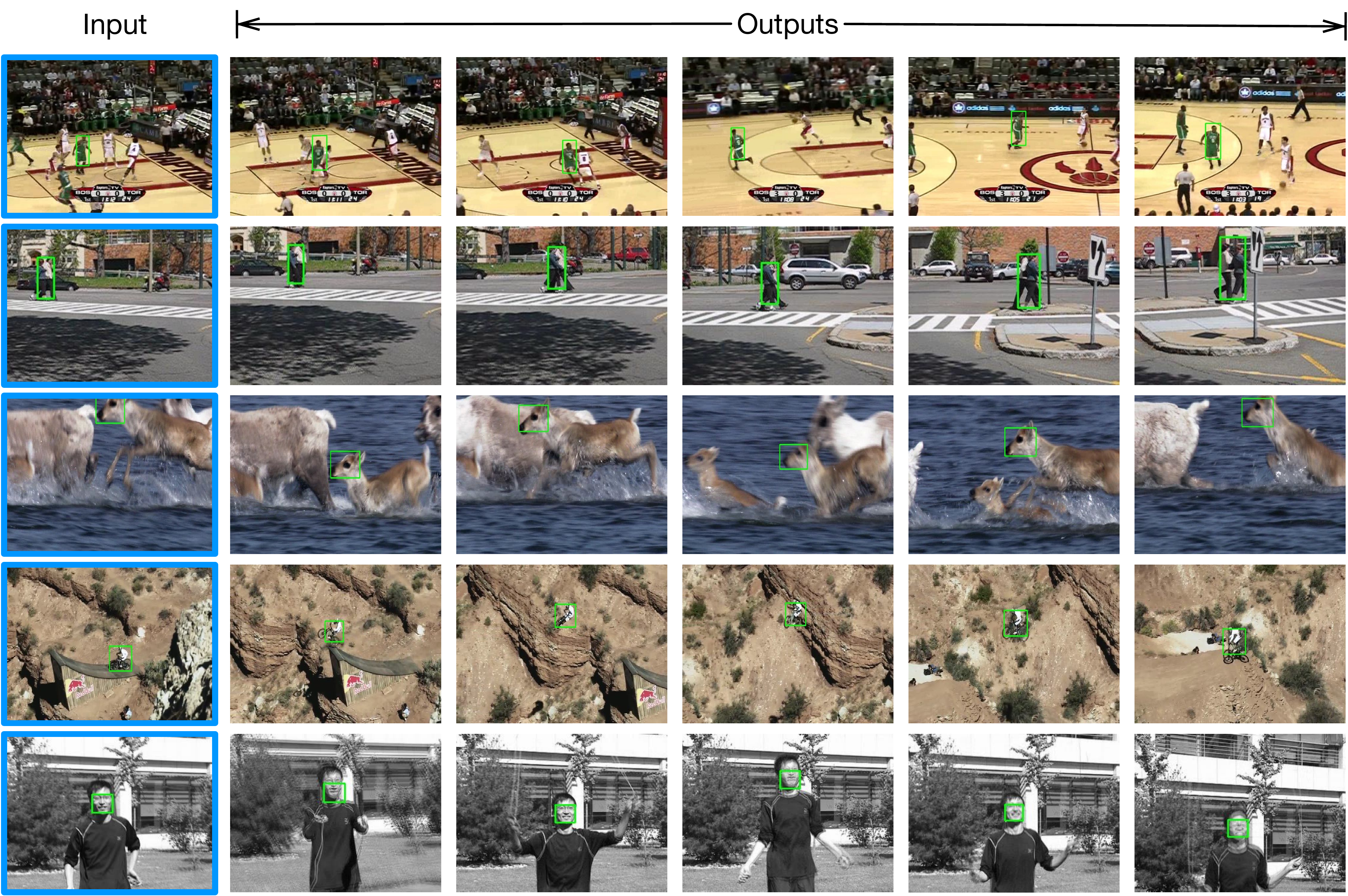}
\vspace{.2em}
\caption{
Qualitative Results for visual object tracking on OTB-100~\cite{yi2015otb}. 
\label{fig:vis_otb}
\vspace{-1.5em}
}
\end{figure*}

We use fine-tuned res$_5$ features for object-level correspondence on OTB-100  visual object tracking~\cite{yi2015otb}.
The results are visualized in Figure~\ref{fig:vis_otb}.
Our VFS could robustly track the target object even under difficult scenarios. 
For example, in the first row, there are multiple similar basketball players, and tracking target undergoes complicate object interaction as well as occlusion. 
Similarly for the deer in the third row, where the tracking target overtakes other similar deers.
For the jumping person in the last row, the video suffers motion blur and large camera displacement.

We provide more visualization in our \href{https://jerryxu.net/VFS/}{project page} .

\end{document}